# YOLO-PPA based Efficient Traffic Sign Detection for Cruise Control in Autonomous Driving

Jingyu Zhang *

The University of Chicago, Chicago,60637, United States, simonajue@gmail.com

Wenqing Zhang

Washington University, St. Louis, 63130, United States, wenqing.zhang@wustl.edu

Chaoyi Tan

Northeastern University, Boston, 02115, United States, tan.chaoyi@icloud.com

Xiangtian Li

University of California San Diego, San Diego, 95035, United States, xil160@ucsd.edu

Qianyi Sun

Vanderbilt University, Nashville, 37235, United States, qianyijay002@gmail.com

**Abstract**

It is very important to detect traffic signs efficiently and accurately in autonomous driving systems. However, the farther the distance, the smaller the traffic signs. Existing object detection algorithms can hardly detect these small-scaled signs. In addition, the performance of embedded devices on vehicles limits the scale of detection modesl. To address these challenges, a YOLO-PPA based traffic sign detection algorithm is proposed in this paper. Firstly, we introduce Parallelized Patch Aware Attention (PPA) into YOLO. PPA utilizes multi-branch feature extraction strategy and spatial-channel attention to capture features at different scales and levels, thereby enhancing the feature extraction capability for long-distance traffic signs. To improve model efficiency without reducing accuracy, we introduce Partial Convolution (PConv) into the YOLO's C2F module. During model training, we introduce APLoss to replace the original classification loss to tackle the serious imbalance of traffic sign categories. The experimental results on the GTSDB dataset show that compared to the original YOLO, the proposed method improves inference efficiency by 11.2%. The mAP@50 is also improved by 93.2%, which demonstrates the effectiveness of the proposed YOLO-PPA.

CCS CONCEPTS

Computing methodologies~Aritificial intelligence~Computer vision~Computer vision problems~Object detection

# 1 INTRODUCTION

With the rapid development of autonomous driving, traffic sign detection has become a key topic in autonomous vehicle cruise control systems. Traffic signs provide important road condition information for autonomous vehicles, including speed limits, driving directions, parking instructions, etc. These information are crucial for vehicle navigation and driving decisions. Accurate traffic signs detection in complex road environments can improve driving safety and efficiency significantly, thereby avoiding traffic accidents. However, due to the diversity and of traffic signs and the complex road environment, as well as the high-speed dynamic changes of vehicles, the practical applications of traffic sign detection in autonomous driving systems face multiple challenges[1].

Most traditional detection algorithms perform well on large objects and simple backgrounds, but their accuracy and robustness often decreases when dealing with small traffic signs and complex backgrounds. Firstly, traffic signs are usually small in size, especially from a far distance. These signs only occupy a small number of pixels in the image and are easily overlooked or misdetected. In addition, traffic signs are often deployed in complex backgrounds. Interference such as trees, buildings, and other traffic facilities, further increases the difficulty of detection.

In recent years, deep learning based object detection algorithms have been widely applied in various fields [2][3][4][5], and significant progress has been made in the field of traffic sign detection. Deep learning models such as YOLO[6] (You Only Look Once), Faster R-CNN, and SSD (Single Shot MultiBox Detector) are widely used in traffic sign detection tasks due to their superior accuracy and real-time performance. These models extract features directly from images in an end-to-end manner and achieve efficient traffic signs detection. However, although these models have achieved good detection accuracy on many publicly available datasets, there are still some challenges in practical applications. The shapes, colors, sizes, and contents of different traffic signs vary a lot. The traffic signs detection system need to have strong generalization ability to detect different types of signs. Because of the drawbacks of traditional convolutional neural networks[7] (CNNs) in fine-grained feature extraction, these models often perform poorly in detecting small-sized and long-distance traffic signs. Meanwhile, the embedded computing devices equipped in autonomous vehicles have limited performance. Many deep learning models have high detection accuracy. However, due to their complex structure and large number of parameters, these models are difficult to perform efficiently in resource limited embedded systems. Researchers have paid attention to improve detection performance with model compression and feature enhancement. However, maintaining detection accuracy while improving efficiency remains a challenging task. Therefore, how to design an efficient and accurate traffic sign detection model that can be achieved with limited computing resources has become an urgent problem to be solved in the field of autonomous driving. Therefore, it is an urgent problem to design a model that can efficiently and accurately detect traffic signs with limited computing resources in autonomous driving.

To tackle these issues, we propose an efficient traffic sign detection algorithm YOLO-PPA in this paper. We introduce a Parallelized Patch Aware Attention (PPA[8]) mechanism in the traditional YOLO architecture. PPA adopts a multi-branch feature extraction strategy and effectively enhances the feature extraction capability for long-distance traffic signs. In addition, Partial Convolution (PConv[9]) is introduced in the original C2F module to reduce the number of model parameters and improve the inference speed without reducing model accuracy. In addition, to address the issue of uneven category distribution of traffic signs, we adopt the APLoss[10] during model training to improve the classification accuracy of the model. We conducted comparative experiments and ablation analysis on the GTSDB dataset. The experimental results showed that the proposed YOLO-PPA achieves an accuracy of 93.2% mAP@0.5 . Compared to the original YOLOv8n, the accuracy has increased by 7.5%, and the inference speed has increased by 11.2%. Comprehensive experimental analysis demonstrated the effectiveness of the proposed method. The main contributions of this paper can be summarized as follows:

(1) We have optimized YOLOv8n and proposed a new lightweight traffic sign detection model YOLO-PPA. YOLO-PPA introduces Partial Convolution (PConv) in the original feature fusion module C2F, which improves the inference efficiency of the model without loss of accuracy. In addition, we have introduced Parallelized Patch Aware Attention (PPA) mechanism. By improving the multi-branch feature fusion strategy, the feature extraction ability of long-distance traffic signs is enhanced.

(2) To address the issue of uneven distribution of traffic sign categories, we introduced APLoss on the classification branch of detection heads. We optimize the classification task as a sorting task of category label confidence. APLoss models the relationships between samples explicitly, thereby reducing the sensitivity of positive and negative sample ratios.

(3) We conducted comprehensive comparative experiments and ablation analysis on the GTSDB dataset. The experimental results show that the proposed YOLO-PPA outperforms the original YOLOv8n in terms of inference speed, model parameter quantity, and detection accuracy. The improvements demonstrated the effectiveness and practicality of YOLO-PPA.

The remaining parts of this paper are organized as follows: Section 2 is the methodology section of this paper. This section provides a detailed description of the proposed methods, including the overall structure of YOLO-PPA, the mechanism of PPA, the structure of the PConv efficient C2F module, and implementation of APLoss. Section 4 presents experimental results and analysis. We compared the performance of different models on the GTSDB dataset and discussed the improvement of detection performance. Section 5 summarizes the research work of this paper and proposes future research directions.

## 2 METHODOLOGY

### 2.1 Overall Structure

As illustrated in Figure 1, the proposed YOLO-PPA is an improved version of YOLOv8n. YOLP-PPA can be divided into four parts: Input, Backbone, Neck, and Head. The input of the model is resized three channel images with shape $640 \times 640 \times 3$.

The backbone part is the core of model, which is used to extract features from images. The backbone mainly includes multiple convolution and down-sampling layers. Residual connections and bottleneck structures are also introduced to reduce network size and improve performance. In the traditional YOLOv8, the C2F module is the basic module. To reduce the number of model parameters and improve inference efficiency without reduce accuracy, we replaced the Conv2d_3×3 in the C2F module with Partial Convolution (PConv).

The neck part is located between the backbone and the prediction heads. It enhances the feature representations by fusing feature maps from different stages of backbone. In the traditional YOLOv8, there was only a Spatial Pyramid Pooling Fast (SPPF) module for multi-scale feature fusion. SPPF concatenates feature maps with different scales by pooling at different scales, thereby improving the detection capability of objects at different scales. We further introduce Parallelized Patch Aware Attention (PPA) mechanism after SPPF. The feature extraction ability for long-distance small-sized traffic signs are further enhanced by extracting features of different scales and levels with parallel branches.

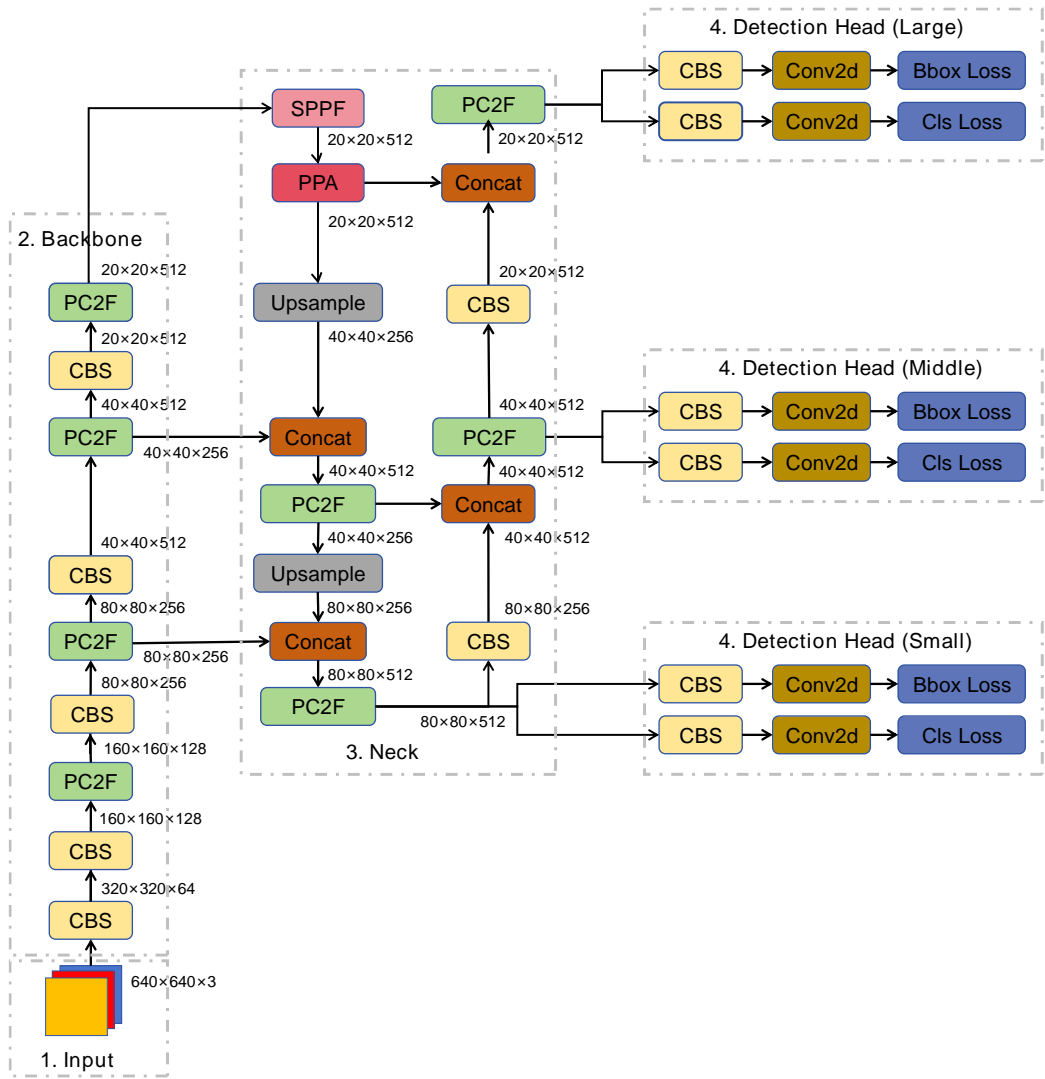

Figure 1. Overall structure of YOLO-PPA.

The head part is responsible for the final object detection and classification. There are totally three heads for detecting objects with three different sizes in YOLOv8. The loss of YOLOv8 includes two parts: the bounding boxes regression loss (CIoU+DFL) and category classification loss (VFL). Due to the significant difference in the amount of different categories in traffic sign detection, we used APLoss during training to improve the classification accuracy of the model.

**2.2 Partial Convolution based Efficient C2F Module**

The C2F modules are the primary feature extraction and fusion components in YOLOv8. Each C2F module consists of multiple bottlenecks. It combines standard convolution and skip connections, which effectively extracts multi-scale features while preserving fine-grained information. The C2F module decomposes the input feature map into several sub-blocks. It progressively constructs multi-level and multi-resolution representations by extracting and fusiong features. However, in complex scenes, C2F may introduce redundant features, leading to decreased computational efficiency. This limitation is particularly evident in embedded systems for automotive vehicles or resource-constrained environments.

To address this issue, we replace the convolutional layers in C2F with Partial Convolution (PConv) and proposed PC2F. In PC2F, multiple Bottlenecks of C2F are replaced with FasterBlock as shown in Figure 2. The Faster Block module is designed with Partial Convolution (PConv) and residual connections. Firstly, the input features are input into a 3×3 PConv layer to extract local information. The features are subsequently further processed through 1×1 convolution, batch normalization (BN), and ReLU activation function. Next, another 1×1 convolution is applied for feature fusion. In the end, the input and processed output are added through residual connections, which preserves the original information and improves computational efficiency.

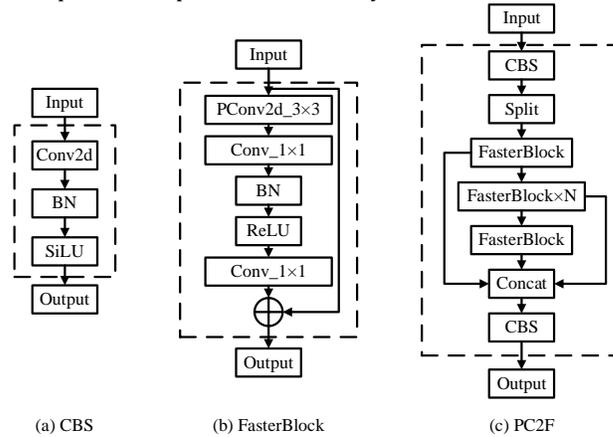

Figure 2. Structure of PC2F and the components in PC2F.

### 2.3 Parallelized Patch Aware Attention

In autonomous driving scenarios, long-distance traffic signs typically have smaller sizes. There are limitations in capturing fine-grained features of small signs in traditional Convolutional Neural Networks (CNNs). Especially in detection models such as YOLO, the limitations of receptive field can easily neglect small-sized objects. [11] Therefore, to improve the detection capability of the model for distant and small signs, we introduce a Parallelized Patch Aware Attention (PPA) mechanism in the neck of YOLO.

The overall structure of the PPA mechanism is presented in Figure 3 (a), which mainly consists of three parallel branches. Each branch is responsible for extracting features of different scales and levels. Specifically, the three parallel branches of PPA include: local branch, global branch, and serial convolution branch. Specifically, local and global branches are constructed by controlling the patch size parameter P.

The input features are first compressed by Point Wise Convolution (PWConv) and input into two patch aware branches with different scales (P=2 and P=4). These paths split the feature map into local patches. The features of these paths are extracted and fused separately. The Patch Aware module in each path performs dimension transformation, pooling, linear transformation, and Softmax weighting on local patches to extract key features. Subsequently, the outputs of the three branches are fused with the CS-Attention module, which enhances key features with Channel Attention and Spatial Attention mechanisms.

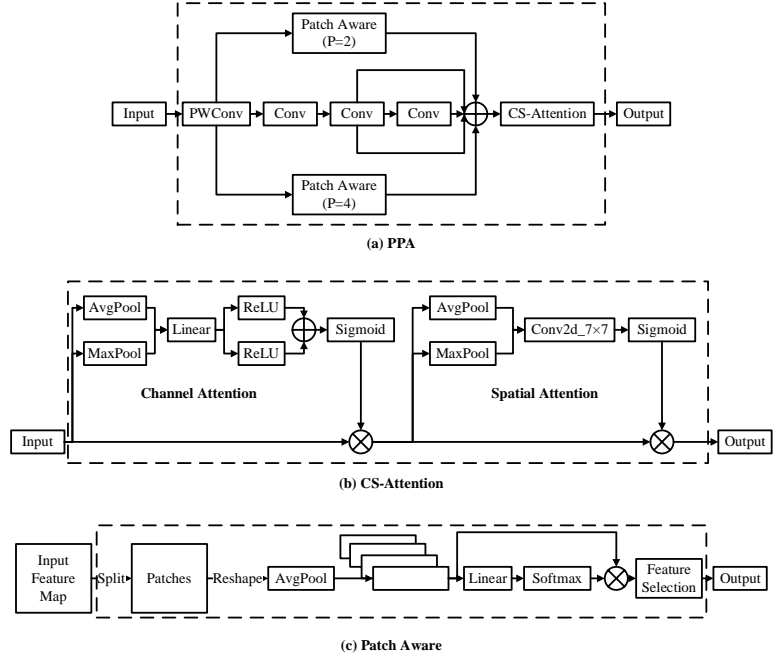

Figure 3. Structure of PPA and its components.

## 2.4 APLoss

There is a significant difference in the number of signs of different categories in traffic sign detection. As illustrated in Equation (1), the original YOLO adopts Cross Entropy loss ($L_{CE}$) for classification.

$$L_{CE} = -\sum_{i=1}^{C} t_i \cdot \log(p_i) \tag{1}$$

Here, $C$ is the number of categories, $t_i$ represents the true label of category $i$ (one-hot encoding), and $p_i$ is the probability of category $i$ predicted by the model. Cross Entropy loss minimizes the distance between the predicted probability distribution and the true distribution, making the model to gradually converge to more accurate classification results during training. [12] But when the categories distribution is extreme uneven, Cross Entropy loss often leads to better performance on frequent categories than rare categories.

To address this issue, we adopt APLoss as the classification loss during the training process as illustrated in Equation (2). APLoss sorts the category prediction results and focuses the model's attention on rare categories. The core of APLoss is to adopt Average Precision (AP) as optimization objective. APloss directly optimizes the evaluation criteria, theoretically enabling more accurate adjustment of model predictions and improving the balance between recall and accuracy.

$$L_{AP} = \frac{1}{|P|}\sum_{i\in P}\sum_{j\in N} L_{ij} = \frac{1}{|P|}\sum_{i\in P} \frac{\sum_{j\in N} H(x_{ij})}{1+\sum_{k\in P\cup N, k\neq i} H(x_{ik})} \tag{2}$$

The molecule of $L_{AP}$ represents the ranking loss between positive and negative samples $i$ and $j$. If the score of negative sample $j$ is higher than that of positive sample $i$, $H$=1; otherwise, it is 0; The denominator represents the correct ranking of sample $i$ among all samples, which is a constant value for sample $i$. The more negative samples with scores higher than positive sample $i$, the larger the $L_{AP}$.

## 3 EXPERIMENTS

### 3.1 Experimental Settings

We used the GTSDB dataset for model evaluation in the experiments. GTSDB contains 900 images with 42 subcategories of traffic signs in 4 major categories (prohibited, dangerous, mandatory, other), and covers traffic signs of various sizes and distances. The experiments were performed on NVIDIA RTX 3090 GPU, with software environments of Ubuntu 20.04, Python 3.8, and PyTorch 1.12. The training settings include a learning rate of 0.001, a batch size of 16, 100 training epochs, an optimizer of Adam, and a cosine annealing learning rate scheduling strategy for optimization. By comparing the performance of the model under different configurations, we evaluated the detection accuracy, inference speed, and model parameter count to validate the effectiveness of the proposed method. The evaluation indicators include Precision, Recall, mAP@0.5, FLOPs, inference speed and model parameter amount.

### 3.2 Comparative Experiments

Table 1 shows the comparative experimental results, where YOLO-PPA presents significant improvements compared to YOLOv8n in all evaluation metrics. Specifically, YOLO-PPA achieves Precision and Recall rates of 91.2% and 89.4%, respectively, far exceeding traditional YOLOv8n's 85.6% and 81.2%. In addition, the mAP@0.5 YOLO-PPA is 93.2%, which is 7.5% higher than YOLOv8n's 85.7%. YOLO-PPA further demonstrates superiority in terms of inference speed and model parameter amounts. [13] The inference speed of YOLO-PPA reaches 145.4 FPS, which is significantly improved compared to YOLOv8n's 130.8 FPS. The model parameter count is lower, only 2.7M. The comparative experimental results indicates that the detection performance has been significantly enhanced in small objects and complex scenes.

Table 1. Comparative experimental results.

| Model | Precision | Recall | mAP@0.5 | FPS | Parameters(M) |
|---|---|---|---|---|---|
| YOLOv5s | 83.5 | 81.0 | 84.4 | 102.1 | 9.1 |
| YOLOv6n | 79.3 | 76.9 | 81.6 | 128.7 | 4.3 |
| YOLOv7-tiny | 85.1 | 82.6 | 86.0 | 122.6 | 6.0 |
| YOLOv8s | 86.3 | 84.5 | 87.1 | 95.2 | 11.2 |
| YOLOv8n | 85.6 | 81.2 | 85.7 | 130.8 | 3.1 |
| YOLO-PPA | 91.2 | 89.4 | 93.2 | 145.4 | 2.7 |

### 3.3 Ablation Study

As shown in Table 2, to further discuss the performance of PPA mechanism and APLoss, we further conducted ablation experiments. The results of the ablation experiment indicate that the PPA mechanism and APloss play a key role in improving the model performance. When only the PPA mechanism is introduced, mAP@0.5 reached 90.4%, demonstrating the effectiveness of PPA in enhancing feature extraction capability. When only introducing APLoss, the improvement of mAP@0.5 reached 3.2%. When both PPA and AP Loss are introduced simultaneously, the model performance reaches its optimal level and mAP@0.5 is improved to 93.2%. The ablation studies demonstrate the effectiveness of these two modules.

ACM-8Table 2. Ablation studies.

| PPA | APLoss | Precision | Recall | mAP@0.5 |
|-----|--------|-----------|--------|---------|
| × | × | 84.5 | 83.8 | 86.2 |
| √ | × | 89.2 | 86.4 | 90.4 |
| × | √ | 87.9 | 86.9 | 89.5 |
| √ | √ | 91.2 | 89.4 | 93.2 |

## 4 CONCLUSION

We propose a YOLO-PPA based efficient traffic sign detection method in this paper. YOLO-PPA effectively improves the detection performance of long-distance and small-sized traffic signs through Parallelized Patch Aware (PPA) attention mechanism, Partial Convolution, and APLoss. The experimental results demonstrate that YOLO-PPA significantly reduces the demand for computing resources while maintaining model accuracy. In the future, we will focus on multi-task learning to balance the collaborative development of traffic sign detection and other autonomous driving tasks.

## REFERENCES

[1] Li S, Mo Y, Li Z. Automated pneumonia detection in chest x-ray images using deep learning model[J]. Innovations in Applied Engineering and Technology, 2022: 1-6.

[2] Bo S, Zhang Y, Huang J, et al. Attention Mechanism and Context Modeling System for Text Mining Machine Translation[J]. arXiv preprint arXiv:2408.04216, 2024.

[3] Ao Xiang, Bingjie Huang, Xinyu Guo, Haowei Yang, and Tianyao Zheng. 2024. A neural matrix decomposition recommender system model based on the multimodal large language model. arXiv preprint arXiv:2407.08942.

[4] Danqing Ma, Meng Wang, Ao Xiang, Zongqing Qi, and Qin Yang. 2024. Transformer-based classification outcome prediction for multimodal stroke treatment. arXiv preprint arXiv:2404.12634.

[5] He S, Zhu Y, Dong Y, et al. Lidar and Monocular Sensor Fusion Depth Estimation[J]. Applied Science and Engineering Journal for Advanced Research, 2024, 3(3): 20-26.

[6] Mo Y, Tan C, Wang C, et al. Make Scale Invariant Feature Transform "Fly" with CUDA[J]. International Journal of Engineering and Management Research, 2024, 14(3): 38-45.

[7] Zongqing Qi, Danqing Ma, Jingyu Xu, Ao Xiang, and Hedi Qu. 2024. Improved YOLOv5 based on attention mechanism and FasterNet for foreign object detection on railway and airway tracks. arXiv preprint arXiv:2403.08499.

[8] Dai S, Dai J, Zhong Y, et al. The cloud-based design of unmanned constant temperature food delivery trolley in the context of artificial intelligence[J]. Journal of Computer Technology and Applied Mathematics, 2024, 1(1): 6-12.

[9] Wang Z, Yan H, Wei C, et al. Research on Autonomous Driving Decision-making Strategies based Deep Reinforcement Learning[J]. arXiv preprint arXiv:2408.03084, 2024.

[10] Gao H, Wang H, Feng Z, et al. A novel texture extraction method for the sedimentary structures' classification of petroleum imaging logging[C]//Pattern Recognition: 7th Chinese Conference, CCPR 2016, Chengdu, China, November 5-7, 2016, Proceedings, Part II 7. Springer Singapore, 2016: 161-172.

[11] Wang X S, Moore S A, Turner J D, et al. A model-free sampling method for basins of attraction using hybrid active learning (HAL)[J]. Communications in Nonlinear Science and Numerical Simulation, 2022, 112: 106551.

[12] Tan C, Wang C, Lin Z, et al. Editable Neural Radiance Fields Convert 2D to 3D Furniture Texture[J]. International Journal of Engineering and Management Research, 2024, 14(3): 62-65.

[13] Wang X. Nonlinear Energy Harvesting with Tools from Machine Learning[D]. Duke University, 2020.